\documentclass[10pt,twocolumn,letterpaper]{article}
\usepackage[accsupp]{axessibility}
\usepackage{iccv}
\usepackage{times}
\usepackage{epsfig}
\usepackage{graphicx}
\usepackage{amsmath}
\usepackage{amssymb}
\usepackage{multirow}
\usepackage{booktabs}
\usepackage{makecell}
\usepackage{xcolor}

\usepackage{authblk}

\usepackage[pagebackref=true,breaklinks=true,letterpaper=true,colorlinks,bookmarks=false]{hyperref}

\iccvfinalcopy % *** Uncomment this line for the final submission

\def\iccvPaperID{6194} % *** Enter the ICCV Paper ID here

\ificcvfinal\fi

\begin{document}

%%%%%%%%% MICROs
\global\long\def\th{\theta}%
\global\long\def\x{\textbf{x}}%
\global\long\def\R{\mathbb{R}}%
\global\long\def\E{\mathbb{E}}%
\global\long\def\l{\ell}%
\global\long\def\k{\mathsf{K}}%
\global\long\def\ste{\text{STE\_Mask}}%
\global\long\def\a{\text{SA}}%
\global\long\def\wa{\text{WSA}}%
\global\long\def\bev{\text{BEV}}%
\global\long\def\h{h}%
\global\long\def\L{\mathcal{L}}%
\global\long\def\hm{m}%
\global\long\def\1{\textbf{1}}%
\global\long\def\hf{\hat{f}}%

%%%%%%%%% TITLE
\title{Efficient Transformer-based 3D Object Detection with Dynamic Token Halting}

% \author[1,2]{Mao Ye}
\author[1,2]{Mao Ye\thanks{mao.ye@getcruise.com}}
\author[2]{Gregory P. Meyer\thanks{greg.meyer@getcruise.com}}
% \author[2]{Yuning Chai\thanks{yuning.chai@getcruise.com}}
% \author[1]{Qiang Liu\thanks{lqiang@cs.utexas.edu}}
\author[2]{Yuning Chai}
\author[1]{Qiang Liu}
\affil[1]{The University of Texas at Austin}
\affil[2]{Cruise LLC}

% \author{Mao Ye\thanks{Work done while intern at Cruise.}\\
% UT Austin\\
% {\tt\small maoye21@utexas.edu}
% \and
% Gregory P. Meyer\\
% Cruise\\
% {\tt\small greg.meyer@getcruise.com}
% \and
% Yuning Chai\\
% Cruise\\
% {\tt\small yuning.chai@getcruise.com}
% \and
% Qiang Liu\\
% UT Austin\\
% {\tt\small lqiang@cs.utexas.edu}
% }

\maketitle
\ificcvfinal\thispagestyle{empty}\fi

%%%%%%%%% ABSTRACT
\begin{abstract}
    Balancing efficiency and accuracy is a long-standing problem for deploying deep learning models. The trade-off is even more important for real-time safety-critical systems like autonomous vehicles. In this paper, we propose an effective approach for accelerating transformer-based 3D object detectors by dynamically halting tokens at different layers depending on their contribution to the detection task. Although halting a token is a non-differentiable operation, our method allows for differentiable end-to-end learning by leveraging an equivalent differentiable forward-pass. Furthermore, our framework allows halted tokens to be reused to inform the model's predictions through a straightforward token recycling mechanism. Our method significantly improves the Pareto frontier of efficiency versus accuracy when compared with the existing approaches. By halting tokens and increasing model capacity, we are able to improve the baseline model's performance without increasing the model's latency on the Waymo Open Dataset.
\end{abstract}

%%%%%%%%% BODY TEXT

\section{Introduction}
Recent progress has shown the potential of applying the transformer architecture, which has been widely used by the natural language processing community \cite{vaswani2017attention,kenton2019bert}, to computer vision tasks
\cite{dosovitskiy2021an,liu2021swin}. Transformers already meet or exceed the performance of convolutional neural networks. However, transformer architectures often suffer from high latency, which is crucial for real-time safety-critical or edge-computing applications.

This paper explores how to speed-up transformer-based 3D object detection. Our method is inspired by network pruning, which increases efficiency by removing less important parts of the model. However, instead of pruning neurons like in \cite{liu2017learning,liu2018rethinking,ye2020good}, our approach prunes or halts tokens. We want to reduce superfluous tokens because the computational complexity of the transformer's attention mechanism increases with the number of tokens. Unlike network pruning, deciding whether or not to halt a token needs to depend on the input, since the importance of a token will depend on the particular example.

% A common practice for accelerating CNN is to prune less important parts of the model. As pruning a certain part of pixels of the latent feature map does not really improve the efficiency of the convolution operations in GPU, a large portion of the current works explore filter pruning \cite{liu2017learning,liu2018rethinking,ye2020good}. Different from CNN, pruning the pixel (i.e, token or patch) of the feature map in the self-attention seems to be more efficient as the computational complexity of self-attention grows quadratically w.r.t. the number of tokens. Different from filter pruning, token pruning is data-dependent as the importance of tokens at different positions varies in different inputs. Recently, several works explore different ways to learn token halting modules to progressively halt tokens in the feed-forward computation in image classification.

There are several recent works that attempt to dynamically halt tokens throughout a vision transformer \cite{rao2021dynamicvit, yin2022vit,pan2021ia, fayyaz2021ats, liang2022evit, tang2022patch}. However, these works consider the task of image classification, while we focus on 3D object detection. Going from image classification to 3D object detection has its challenges, but also its benefits.
A challenge that arises with object detection is that any token could contain an object; therefore, we require a method that can detect objects from all tokens regardless of whether or not they were halted. This issue does not occur for image classification with vision transformers. For image classification, an artificial token is often added to classify the image, and this token is prevented from being halted.
Since the halting of a token is a non-differentiable operation, a new design of the computation graph is needed to define pseudo-gradients that enable the back-propagation during training.
A benefit with object detection is that the labels contain not only the object's classification but also its location, and our approach is able to leverage the localization of the objects to help the model learn which tokens are more important than others. 

Our method is designed for 3D object detection for autonomous driving. For safety-critical tasks, it is often important that the model is non-stochastic. Therefore, unlike \cite{rao2021dynamicvit,fayyaz2021ats}, our halting module is designed to be deterministic.

% The goal of this work is to explore how to dynamically halt the token for the transformer-based 3D detector. Several improvements and adaptations need to be made to train an effective token halting module for the more complicated task and network architecture: 1. 3D detection network uses a different mechanism for inference such as bird-eye view (BEV) rather than a $\mathtt{cls}$ token in image classification. To construct an informative BEV, we need to aggregate the information in both the early halted and not halted tokens; 2. Such an architecture difference also requires a new design of the training computation graph to enable the backpropagation for training the halting modules; 3. the training set of 3D detection provides a more fine-grained bounding box label, such label enables us to train the halting module in a more effective supervised manner rather than the purely unsupervised approach used in the existing works; 4. we consider the self-driving application that requires the network to be deterministic during the inference, which means the output of the halting module needs to be non-stochastic. Such constrain bring extra optimization difficulty for the existing probabilistic-based framework.

Our contributions can be summarized as follows:
\begin{itemize}
    \itemsep0em 
    \item We propose a deterministic module that progressively halts tokens throughout the transformer, and a simple but effective token recycling mechanism to reuse the information from the halted tokens.
    \item An equivalent differentiable forward-pass is proposed to overcome the non-differentiability of token halting, and a theoretical analysis is conducted to evaluate the accuracy of the pseudo-gradient.
    \item A non-uniform token sparsity loss is employed to improve the learning of the halting module by utilizing the ground-truth bounding boxes.
\end{itemize}
\section{Related Work}
\subsection{Dynamic Transformer}
The idea of adapting the number of tokens within a transformer to improve performance has recently been explored. \cite{rao2021dynamicvit, yin2022vit} learn a token selection module to dynamically halt tokens at inference using the Gumbel-softmax trick \cite{jang2016categorical} and ACT \cite{graves2016adaptive}, respectively. \cite{pan2021ia, fayyaz2021ats, liang2022evit, tang2022patch} propose different heuristics based on the attention weights to halt or aggregate tokens. \cite{kong2021spvit} combines both token selection and aggregation. \cite{xu2022evo} proposes a slow-fast token update that applies token-wise transformations on the halted tokens and attention-based transformations on those that are not halted. \cite{zong2022selfslimming} proposes to globally aggregate the tokens into a smaller set of new tokens using a reconstruction loss, which encourages the new tokens to preserve as much information as possible. Instead of adaptively removing or combining tokens, \cite{wang2021not,zhu2021make} consider using tokens with adaptive spatial size for different inputs, and \cite{meng2022adavit} simultaneously selects the tokens, attention heads, and attention windows.

The majority of the existing works are designed for image classification, while we consider the task of 3D object detection. Consequently, the prior work cannot be directly applied to object detection and significant algorithmic changes are required. Vision transformers for image classification use an artificial token, typically referred to as the $\mathtt{cls}$ token, to perform classification. However, for object detection, all tokens are used by a detection head. Therefore, our method requires a way to aggregate information from all tokens regardless of if or when a token was halted. Additionally, the aggregation of halted tokens needs to be differentiable. As a result, our proposed method differs significantly from the previous work.

Two prior works that do consider the task of object detection are \cite{roh2022sparse} and \cite{zhang2022not}. \cite{roh2022sparse} applies dynamic token selection within the Deformable DETR framework \cite{zhu2021deformable}, and \cite{zhang2022not} employs non-uniform point cloud downsampling for a point-based transformer. We consider \cite{roh2022sparse} orthogonal to our approach as it focuses on improving the convergence of the transformer detection head, while we focus on improving the backbone. Furthermore, Deformable DETR selects a fixed subset of keys/values for each query. We select a dynamic subset of tokens, which in turn reduces all queries, keys, and values. In our case, the token selection is more challenging because we need to determine the token importance in the early stages of the backbone, where the token features are less informative. However, Deformable DETR is applied to the detection head and has access to deep features extracted from the backbone. In terms of  \cite{zhang2022not}, it attempts to downsample the background points, while our approach tries to halt any unnecessary tokens. Due to the different goals, the algorithms are considerably different.

\subsection{Efficient Network Architecture}
Trading off between network efficiency and accuracy has been a long-standing problem. Common approaches such as network pruning \cite{liu2017learning,frankle2018lottery,liu2018rethinking,ye2020good,ye2020greedy,tanaka2020pruning,ye2020greedy,yu2021unified}, network quantization \cite{rastegari2016xnor,zhou2016dorefa,zhao2019improving,nagel2019data,banner2019post,liu2021post}, and neural architecture search \cite{sandler2018mobilenetv2,liu2018darts,liu2018progressive,tan2019efficientnet,liu2019auto,cai2019once,gong2022nasvit,li2022efficientformer,chavan2022vision, liu2022lidarnas} have been proposed to push the limit of the Pareto front. Compared with our method, those approaches aim to search for a fixed network architecture that is not adaptive to the input. On the other hand, learning a network with a dynamic computation graph has also been explored in various directions including adaptive resolution \cite{najibi2019autofocus,yang2020resolution,meng2020ar}, depth \cite{figurnov2017spatially,veit2018convolutional,wang2018skipnet,wu2018blockdrop}, and channels \cite{lin2017runtime,yu2019universally,Bejnordi2020Batch-shaping}. Those directions are orthogonal to this work and could be performed in combination with our method.

\subsection{LiDAR-based 3D Object Detection}
Existing networks for 3D detection can be classified based on the representation of the 3D scene. The most popular approach is to represent the scene using a voxel grid. Vote3Deep \cite{engelcke2017vote3deep} was one of the first to use a uniform voxel grid to represent the point cloud. The representation has been further improved upon by \cite{zhou2018voxelnet,yan2018second} using a small PointNet \cite{qi2017pointnet} to learn a better voxel representation, and by \cite{graham20183d} using sparse 3D convolutions to improve the efficiency. The efficiency was improved further by \cite{yang2018pixor,lang2019pointpillars} using 2D convolutions instead of 3D convolutions. Another popular approach is to directly process the point representation \cite{qi2019deep,shi2019pointrcnn,chen2020object,wong2020identifying}. These methods are usually built on PointNet~\cite{qi2017pointnet} and require a nearest neighbor search between points which can be difficult to scale. Finally, there are a handful of methods that represent the 3D scene using a range image \cite{meyer2019lasernet,liang2020rangercnn,meyer2020laserflow,meyer2020learning,fan2021rangedet,chai2021point}.

Due to the recent progress in transformers for computer vision \cite{dosovitskiy2021an}, the transformer architecture has also been applied to 3D object detection. Existing works include transformer-based backbones \cite{mao2021voxel,he2022voxel,fan2022embracing,sun2022swformer,fan2022fully, wang2023dsvt} for the voxel-based representation, \cite{misra2021end,zhang2022not} for the point-based representation, and \cite{guan2022m3detr} for a combination of  both the point and voxel representation. Furthermore, transformers have been used to improve the detection head \cite{zhou2022centerformer} and for sensor fusion \cite{bai2022transfusion,zhang2022cat,wang2022bridged}. 

Similar to our work, \cite{sun2021rsn} proposes to use a foreground point selection to remove LiDAR points that do not belong to objects. This idea was later applied to a transformer-based detector \cite{sun2022swformer}. This selection process is based purely on whether the points/voxels belong to foreground objects. In comparison, our dynamic token halting is learned based on a token's contribution to the detection task. That is, important background tokens can be kept while less important foreground tokens can also be removed. In addition, our framework incrementally halts tokens, and all tokens are used to inform the final predictions, while in \cite{sun2021rsn, sun2022swformer} voxels/points are simply removed early in the network. Lastly, the voxel selection in \cite{sun2022swformer} is applied at the end of feature learning and thus has only a mild impact on latency. In addition, \cite{chen2022focal} proposes to select a subset of voxels as input to sparse convolutions. An important difference between our approach and \cite{chen2022focal} is how non-differentiable operations are handled. We propose an equivalent differentiable forward-pass to enable full differentiability with theoretical support. \cite{chen2022focal} uses a non-differentiable threshold during training; therefore, voxels that are not selected do not receive end-to-end gradient updates.
\begin{figure*}[t]
    \centering
    \includegraphics[width=\textwidth]{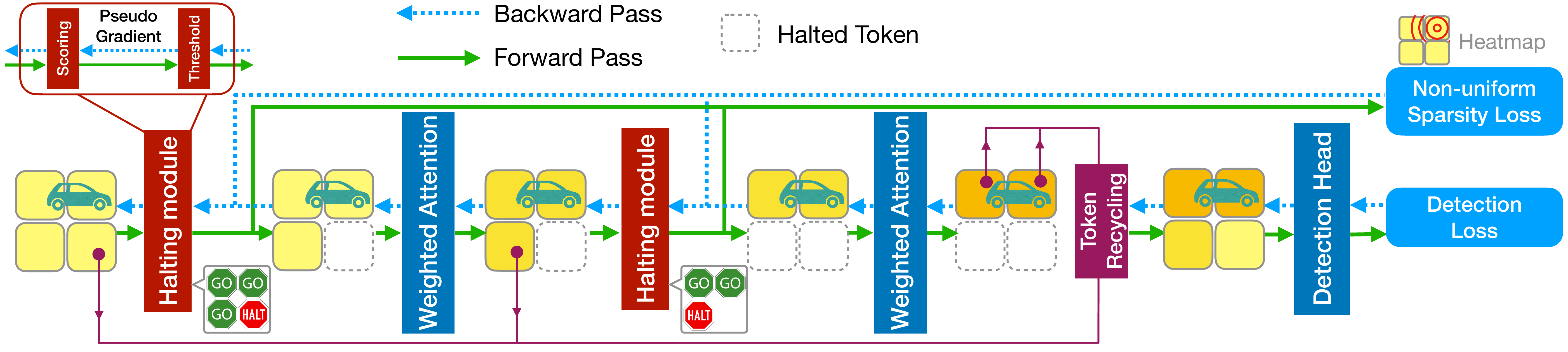}
    \caption{Given a set of tokens, the halting module produces a score for each token, and the tokens with scores below a threshold will be halted. At inference, only tokens that are not halted are forwarded to the next layer. However, during training, all the tokens are forwarded to the next layer, but halted tokens are prevented from interacting with the other tokens. We do this in order to obtain a pseudo-gradient to back-propagate through the non-differentiable threshold operation. After the last attention layer, halted tokens are recycled and combined with the non-halted ones and forwarded to the detection head. The whole network is trained end-to-end using both a detection loss and a non-uniform sparsity loss.}
    % Overall pipeline. Given the features of the not-halted voxels, the halt module produces the voxel scores and voxels with scores below the threshold will be halted. At the training time, the weighted self attention is applied to all the voxels while the weight of the halted voxels are 0 and the weights of the not-halted voxels are their scores. During the inference time, the weighted self attention is only applied to not-halted voxels with weight being their scores. The halting module is trained based on both detection loss and prior sparsity loss. }
    \label{fig:pipeline}
\end{figure*}

\section{Background}
\subsection{3D Object Detection}
Given $P=\{(x,y,z,r,t)_{i}\}$, a point cloud of 3D coordinates $(x,y,z)$, intensity $r$, and elongation $t$ measurements, we want the model to predict a set of 3D bounding boxes for the
objects of interest. The bounding box $b=(l_{x},l_{y},l_{z},w_{x},w_{y},w_{z},\alpha)$
parameterizes the location of an object by its center position
$(l_{x},l_{y},l_{z})$, dimensions $(w_{x},w_{y},w_{z})$,
and heading angle $\alpha$.

\subsection{Transformer-based 3D Object Detector}
Our method is built upon the recent Single-stride Sparse Transformer (SST), which is a LiDAR-based 3D object detector \cite{fan2022embracing}. It uses
dynamic voxelization \cite{zhou2020end} to create a set of voxel features, $f_{0}$, in the bird's eye-view, and each voxel is treated as a token within the transformer. For this reason, we will use the terms voxels and tokens interchangeably. 
SST uses regional grouping to divide the voxel grid into non-overlapping regions, and it applies sparse regional attention (SRA) to the voxels within each region.
Specifically,
% \mao{Seems other vit paper might use $\hat{f}^l$ or $f^{'l}$ for our $f^{l,1}$. Since this notation is only used once, let's keep the current form?}
% \greg{I like $f_l'$ (see comment)}
% \mao{Yeah, originally my l is actually subscript. But it is a bit tricky when we do $f_{l,i}$ in equ (6) when we refer to the $i$-th voxel at layer l.}
% \greg{couldn't you make $i$ a superscript? $f_l^i$?}\mao{I feel $i$ is kind of indexing and more conventional to be a subscript? Like we always all the (i,j)-th element of matrix $M$, $M_{ij}$. If we put the indexing superscript, we need to change all of the later notations.}
\begin{equation}
\text{SRA}(f_{l})=\text{MLP}(\text{LN}(f_l'))+f_l'
\end{equation}
where
\begin{equation}
f_l' =\text{MSA}(\text{LN}(f_{l}),\text{PE}(f_{l}))+f_{l}\text{,}
\end{equation}
and $\text{LN}$, $\text{MLP}$, $\text{PE}$, $\text{MSA}$ denote
layer normalization \cite{ba2016layer}, token-wise multi-layer perceptron,
position encoding, and multi-head self-attention \cite{vaswani2017attention}, respectively. Since objects can lay within several regions, two SRA operations are performed sequentially where the regional grouping is shifted for one of the operations, similar to the Swin Transformer~\cite{liu2021swin}.
Thus, the $(l+1)$-th features are generated by 
\begin{equation} \label{equ:SSTbasicblock}
    f_{l+1}=\text{SSRA}(\text{SRA}(f_{l})),
\end{equation}
where $\text{SSRA}$ is the shifted sparse region attention. After several rounds of attentions, the bird's eye-view feature map is constructed from the final voxel features, and the feature map is passed to a detection head to predict the object bounding boxes.

\section{Proposed Method}
\paragraph{Notation}
Without loss of generality and to simplify the notation, we assume
that each layer of our transformer backbone processes the tokens as follows:

% \mao{Some other paper just uses `Attention` in the equation for attention. Does the current symbol LG?}
% \greg{The different font looks odds. Maybe we should just use `Attention` if that's what others use.} \mao{well, they do not need to call Attention multiple times and usually use only once when they define the basic blocks}
\begin{equation}
    f_{l+1}=\phi_{2}(\a(\phi_{1}(f_{l})), f_{l}),
\end{equation}
where $\phi_{1}$ and $\phi_{2}$ are non-linear token-wise operations, and $\a$ is some general self-attention mechanism (not necessarily the SRA used by SST). Note that each basic block (\ie~\eqref{equ:SSTbasicblock}) of SST can be viewed as two layers using this notation due to the usage of region shift.

\subsection{Overview}
We are motivated by the idea that different tokens may vary in importance to the detection task. 
Furthermore, a token may be useful at the early stages of the network but less informative at the later stages.
For example, detecting a vehicle on an empty road may require several late stage tokens from the vehicle itself, but only a few early stage tokens from the surrounding road.
Alternatively, detecting a camouflaged pedestrian may require several late stage tokens from both the person and the surrounding environment.
Our goal is to learn a token halting mechanism that identifies which tokens should be kept and forwarded
to the subsequent layer. Given $N$
token features, $f_{l}=\{f_{l,i}:i\in\{1,\ldots,N\}\}$, from the $l$-th layer, the halting module outputs a binary mask, $k_{l}\in\{0,1\}^{N}$, indicating which of the tokens are forwarded to the next layer. Specifically, the subsequent layer's tokens are computed as
\begin{equation}
f_{l+1} =\phi_{2}(\a(\phi_{1}(\hf_{l})),\hf_{l})\text{,}
\end{equation}
where $\hf_{l} =\{f_{l,i}: k_{l,i}=1\}$ are the tokens kept by the halting module.
Refer to Figure \ref{fig:pipeline} for illustration of our proposed method.
% Note
% that such probability should be interpreted as conditional probability
% as it is computed based on the condition that the voxels are not halted
% up to the $l$-th layer. We sample voxels based on the keeping probability
% $k$ and only forward those sampled voxels $\bar{f}_{h}^{l}$ to the
% next layer.

% In general, our frameworks allow the keeping probability be any continuous
% value in $[0,1]$ and produce a stochastic network at inference. However,
% in the autonomous driving application, we want the inference of our
% detection model to be non-stochastic which requires a constrain that
% $k\in\{0,1\}$. In this paper, we consider the case that the network
% being deterministic and thus requires $k$ being either 0 or 1. The
% final forward computation is
% \mao{seems we can simply throw away the bar using $\hf$. Does it looks good?}
% \begin{align}
% f^{l+1} & =\phi_{2}(\a(\phi_{1}(\hf^{l})),\hf^{l}),\\
% \text{where }\hf^{l} & =\{f_{i}^{l}:i\in\{1,...,N\}\land k_{i}^{l}=1\}.
% \end{align}

\subsection{The Halting Module} \label{sec:halt}
The halting module outputs a binary mask $k_{l}$ that determines
whether a token is forwarded to the next layer. To
create such a mask, our halting module first computes a non-negative score $s$ for each token. The architecture used to produce the score could be anything; it could be as simple as a MLP or as complex as a transformer. Afterwards, the mask can be obtained by thresholding the score, $k_l=\psi(s_l)$,  where $\psi(s_l)=\mathbb{I}\{s_l\ge u\}$ and $u$ is a threshold. The threshold, $u$, could be a fixed value or selected dynamically using the distribution of the scores.

Each layer has its own halting module as the distribution of features shifts throughout the network. In practice, we found that it is better to use a more complicated architecture for the halting module during the earlier layers, while a simple architecture is sufficient for for the later layers. Refer to Section \ref{sec:exp_details}, for more details on the specific architecture used by our method.

\subsection{Weighted Self-Attention}
To improve the learning of the halting module, we propose a weighted self-attention operation $\wa(f_{l},s_{l})$, which weights the tokens based on the score $s_l$ produced by the halting module.
The output of the weight self-attention for the $i$-th token is defined as follows:
\begin{equation}
\wa(f_l,s_l)_{i} =\frac{\sum_j\exp(P_{ij})s_{l,j} v_{j}}{\sum_{k}\exp(P_{ik})s_{l,k}}
\end{equation}
where
\begin{equation}
    P =(W_{Q}f_{l})(W_{K}f_{l})/\sqrt{d}\text{,}\qquad v = W_{V}f_{l}\text{,}
\end{equation}
and $W_{Q}$, $W_{K}$, and $W_{V}$ are the queue, key, and value matrices, respectively. 
This definition assumes the standard self-attention operation; however, it could be easily applied to other variants.

By using the token score $s_l$ as the weight, we force the attention given to a particular token to be proportional to that token's score.
Intuitively, this encourages the halting module to increase the score of a token if it plays an important role within the attention mechanism.
Refer to Section~\ref{sec:grad}, for an in-depth analysis of how the weighted self-attention affects the learning of the halting module.

\subsection{Token Recycling}
We observe that even when a token is halted early, its features are still useful to inform the final predictions of the model.
Instead of throwing away the halted tokens, we recycle them by directly forwarding them to the detection head.

Recall that SST constructs a bird's eye-view feature map, $f_{\bev}$, from the output of the transformer, and passes that feature map to the detection head.
Since our method is built upon SST, we also construct a bird's eye-view feature map $\hat{f}_{\bev}$; however, in our case, the feature map is assembled from both the output of the transformer and all the halted tokens using their final features at the layer they are halted. 
Refer to Figure \ref{fig:pipeline} for an illustration of this process.

\subsection{Equivalent Differentiable Forward-Pass}

The token halting operation is non-differentiable, which is an issue
for training with gradient descent. To overcome this issue,
we introduce an equivalent differentiable forward-pass (EDF) during training. The goal of EDF is to produce a forward-pass that generates an equivalent output but is differentiable.

Unlike during inference, where the halted tokens are not forwarded to the subsequent layer, EDF forwards all the tokens but prevents ``halted'' tokens from interacting with other tokens by using the mask as a multiplier on their score.
Specifically,
\begin{equation}
{f}_{l+1} = \phi_{2}(\wa(\phi_{1}({f}_{l}),s_{l} \circ k_{0:l}), {f}_{l})\text{,}
\end{equation}
where
\begin{equation}
    k_{0:l}=k_{0} \circ\ldots\circ k_{l}
\end{equation}
and $\circ$ denotes an element-wise multiply. 
Note that, we set $k_{0}:=\textbf{1}$.
For each token, we use its feature at the layer it was halted to construct $f_{\bev}$. 
Similarly, EDF defines the feature map as
\begin{align}
{f}_{\bev} = \sum_{l=1}^{L+1}(k_{0:l-1}-k_{0:l})\circ{f}_{l}\text{.}
\end{align}
For convenience, we set $k_{0:L+1}:=\textbf{0}$ where $L$ is the total number of layers.
Although inference and training have different forward-passes due to EDF, it is trivial to verify that $f_{\bev}=\hat{f}_{\bev}$.
% that produces the
% same output as we do real voxel halting at inference. Specifically,
% at the training time, we consider 
% \begin{align}
% \hat{f}^{l+1} & = \phi_{2}(\a(\phi_{1}(\hat{f}^{l}),s^{l}\circ k^{l}), \hat{f}^{l})\\
% \hat{f}_{i}^{\bev} & =\sum_{l=1}^{L}(k^{0:l-1}-k^{0:l})\circ\hat{f}^{l},
% \end{align}
% where $\circ$ denotes the element-wise multiply and the $k^{0:l}$ defines the joint keeping mask by multiplying the keeping mask at each layer $k^{0:l}=\prod_{i=0}^{l}k^{i}$
% (here $k^{0}:=\textbf{1}$ and $k^{L}=\textbf{0}$, assuming there are
% $L$ layers in total.). Essentially, $k^{0:l-1}_i-k^{0:l}_i$ is 1 if the $i$-th voxel is firstly halted at layer $l$ else 0 and $\hat{f}_{i}^{\bev}$ is thus the voxel feature at the first layer it is halted. 

% The final BEV map is constructed by $\hat{f}^{\bev}=\{\hat{f}_{i}^{\bev}:i\in[N]\}$.
% The key property is that 
% \begin{equation}
%     \a(\phi_{1}(\hat{f}^{l}),s^{l}\circ k^{l})_{i\cdot}=\begin{cases}
% \a(\phi_{1}(\bar{f}_{h}^{l}),s^{l})_{i\cdot} & \text{if}\ k_{i}^{l}=1\\
% 0 & \text{else}.
% \end{cases}
% \end{equation}
% This means that once a token is halted, the weighted attention zeros out its interaction with other voxels and thus produces the same output as if the token halting is conducted.
% It is not hard to verify $\hat{f}^{\bev}=f^{\bev}$.

The binary mask $k_l$ is produced by thresholding the token's score $s_l$, and this thresholding function $\psi(s_l)$ has zero gradients almost everywhere.
To enable back-propagation, we use the straight-through estimator (STE) \cite{bengio2013estimating}. STE defines a pseudo-gradient during the back-propagation, by replacing the derivative of the threshold function with the derivative of some other activation function
\begin{equation}
    \psi'(s_l):=\sigma'(s_l)\text{.}
\end{equation}
A common choice for $\sigma$ is the identity function.

By combining the EDF with the STE, our computation graph at training is
fully differentiable. In Section \ref{sec:grad}, we provide a detailed
analysis of the (pseudo)-gradients used to update the halting module.

\subsection{Non-Uniform Token Sparsity Loss} \label{sec:prior}
We want to encourage the halting module to output a sparse binary mask, and a common approach \cite{li2016pruning, ye2018variable} is to penalize the scores $s_l$ with a $\ell_1$ penalty (\ie LASSO \cite{tibshirani1996regression}). Such a penalty is applied uniformly to all the tokens. On the other hand, we find that the performance can be significantly improved by applying a non-uniform penalty to the tokens. Our intuition is that tokens belonging to a foreground object are usually more important than ones belong to the background. Therefore, on average, we want foreground tokens to have a larger score.
To accomplish this, we leverage the grouth-truth bounding boxes and create a heatmap in
a similar fashion as \cite{zhou2019objects}. Tokens that are within a bounding box have a positive value in the heatmap between $[0, 1]$, and a token's value increases as it gets closer to the center of object.
Tokens that do not fall inside any bounding box have a value of zero in the heatmap. 
Any difference between $s_l$ and the heatmap is penalized using focal loss \cite{lin2017focal}.
Such a loss applies a uniform sparse penalty to the background token and a non-uniform penalty to the foreground tokens based on their distance to the object's center. See Appendix \ref{apx:non-unif} for more details.

\subsection{Losses}
The feature map $f_{\bev}$ is forwarded to the CenterPoint~\cite{yin2021center} detection head for predictions, and we train the model end-to-end with a total loss defined as
\begin{equation}
    \L = \lambda_b \L_{b} + \lambda_h \L_{h} + \lambda_s \L_{s},
\end{equation}
where $\L_{b}$ and $\L_{h}$ are the box and heatmap regression losses defined by \cite{yin2021center}, and $\L_{s}$ is our non-uniform sparsity loss.
\section{Analyzing the Pseudo-Gradient} \label{sec:grad}
We demonstrate that the pseudo-gradient provided by our proposed EDF and the STE gives a high-quality update direction for the halting module. To simplify the analysis, we consider a reduced model
with only one attention layer. In this case, $f_{\bev}=\{q_{1},q_{2}\}$
where we define
\begin{equation}
    q_{1}=(\1-k)\circ f,\ q_{2}=k\circ\phi_{2}(\wa(\phi_{1}(f),s\circ k),f)\text{.}
\end{equation}
Note that we drop the subscripts to simplify the notation.
Furthermore, we use a STE of $\psi'(s)=1$.
Consider the situation that the $i$-th feature is halted in the first
layer, \ie $k_{i}=0$, while it is actually better to forward
to the next layer. In such case, a
feature map $\tilde{f}_{\bev}=\{\tilde{q}_{1},\tilde{q}_{2}\}$, where
\begin{align}
\tilde{q}_{1} & =(\1-k-\textbf{\ensuremath{\1}}_{i})\circ f\\
\tilde{q}_{2} & =(k+\1_{i})\circ\phi_{2}(\wa(\phi_{1}(f),s\circ(k+\1_{i})),f)\text{,}
\end{align}
and $\1_{i}=[0,0,\ldots,1,0,0,\ldots]$ (only the $i$-th index being 1), is expected to have a smaller loss.
Specifically,
\begin{equation}
    \Delta_{i}:=\L(\tilde{q}_{1},\tilde{q}_{2})-\L(q_{1},q_{2})<0\text{,}
\end{equation}
where $\L$ is a detection-related loss. We are able to show that 
\begin{equation}
    \Delta_{i}\approx \frac{\partial\L(q_{1},q_{2})}{\partial{s_{i}}} + O(u)\text{,}
\end{equation}
where the approximate equality has precision up to the second-order Taylor approximation.
That is to say, our pseudo-gradient is almost identical to the change of loss, and the additional error term has a magnitude proportional to the threshold $u$. Since the threshold is, in general, very small (a practical choice being $u\sim 0.01$), the pseudo-gradient gives an accurate update direction, pushing the halting module towards a direction that gives a better halting decision. The analysis of when the token is forwarded but it is better to halt is similar and thus omitted. See Appendix \ref{apx:gradient} for more details and a derivation.
\section{Experiment}
\subsection{Setup and Implementation Details} \label{sec:exp_details}

\paragraph{Dataset}
We evaluate our method using the Waymo Open Dataset (WOD) \cite{Sun_2020_CVPR} which contains 1,150 sequences where 798 sequences are for training, 202 for validation, and 150 for testing. The entire dataset contains more than 200k frames, and each frame contains a LiDAR point cloud covering a $150m \times 150m$ area.

\paragraph{Backbone}
For our experiments, we use the default backbone configuration of SST \cite{fan2022embracing}. It uses a dynamic voxelization technique similar to PointPillars \cite{lang2019pointpillars} with the LiDAR point cloud as the input. Furthermore, SST uses four consecutive SRA blocks followed by four dense convolutional layers. Each self-attention operation uses 8 heads, 128 input channels, and 256 hidden channels. Each spatial region contains $14 \times 14 \times 1$ voxels, and each voxel has a size of $0.32m \times 0.32m \times 6m$. We use the CenterPoint \cite{yin2021center} one-stage detection head and single frame of input.

\paragraph{Halting Module}
We employ two dynamic halting modules before the first and second SRA blocks. Since we obtain a very high-level of token sparsity after the second halting module with negligible performance drop (see below), adding more modules to further increase sparsity gives limited speed-up. Like mentioned in Section \ref{sec:halt}, we use a complex architecture for the first halting module and simple architecture for the second one. For the first halting, we use a lightweight U-Net \cite{ronneberger2015u} architecture with MobileNetV2 blocks \cite{sandler2018mobilenetv2}, and a linear layer with sigmoid activation to obtain the halting score. For the second halting module, we simply apply a one-layer MLP on the input feature to produce the score.
For both halting modules, the first 32 features of a token are used as input.
To re-use the latent features extracted by the halting module, we fuse the penultimate latent features back into the token features.

The halting threshold $u$ is adjusted to obtain different levels of token sparsity. The threshold is set such that only a certain quantile of tokens are halted. We found that setting the threshold based on the score quantiles instead of a fixed threshold helps stabilize training. We refer readers to Appendix \ref{apx:setup} for more details.

\paragraph{Training}
We train the network for 24 epochs using AdamW optimizer \cite{loshchilov2018decoupled} and a one-cycle learning rate. The learning rate starts at $4\times 10^{-5}$, and increases to $1\times 10^{-3}$ during the first $10\%$ of iterations. Afterwards, the learning rate is annealed with consine decay for the rest of the iterations. We apply standard data augmentation (random flop, rotation, and scaling) during the training. The loss weights are set to $\lambda_b = 2.0$, $\lambda_h = 1.0$, and $\lambda_s=0.5$.

\subsection{Efficiency and Accuracy Trade-off} \label{sec:pareto}
\begin{figure}[t]
    \centering
    \includegraphics[width=0.48\textwidth]{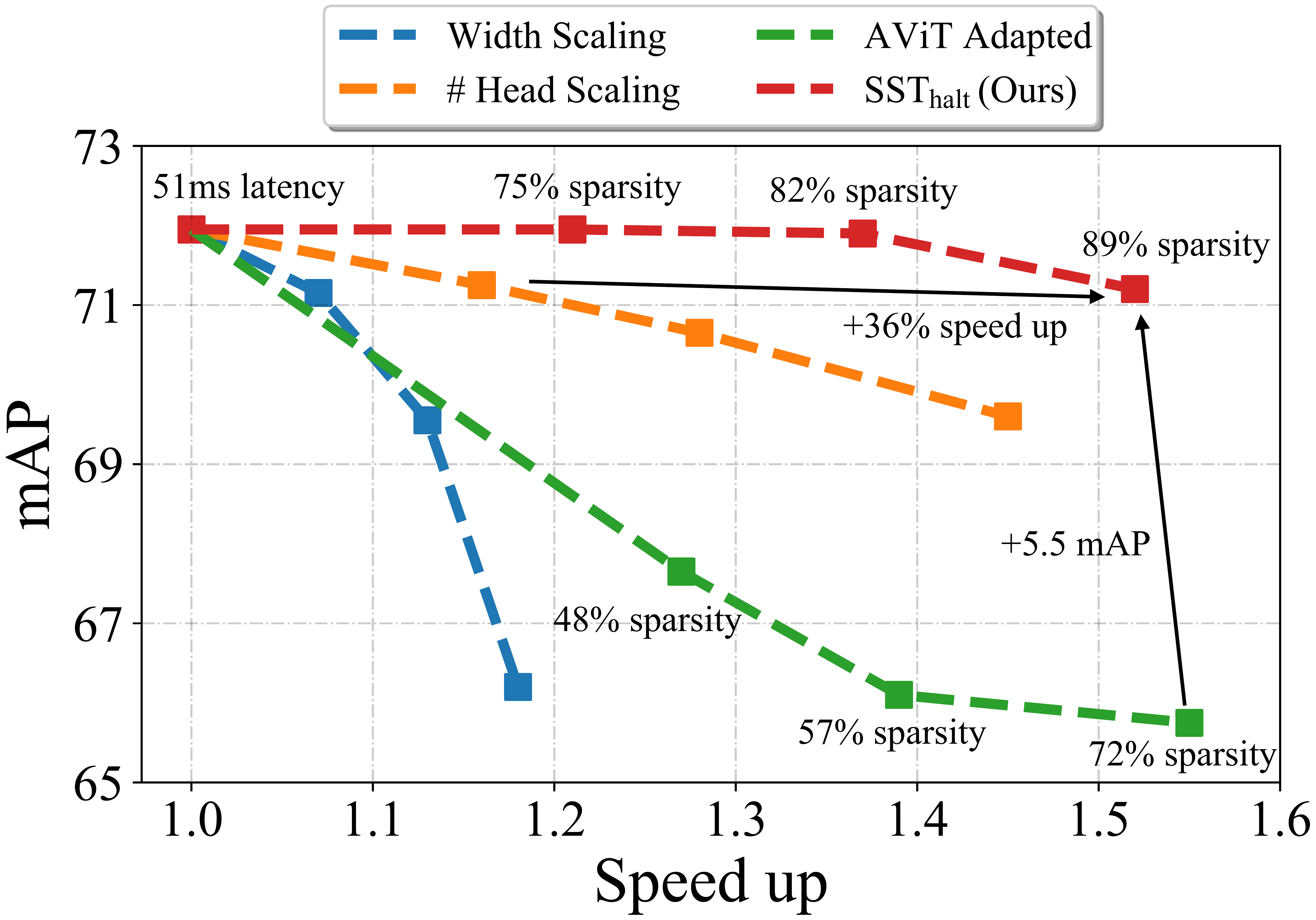}
    \caption{The accuracy-efficiency trade-off. The mAP is the average of L1 and L2 vehicle AP. The sparsity is measured by averaging the percentage of halted tokens across all layers.} 
    \label{fig:pareto}
\end{figure}

The goal of this section is to study the efficiency and accuracy trade-off achieved by varying the sparsity of the tokens. We compare our dynamic token halting approach with other model scaling approaches, including changing the latent dimension of the attention mechanism (\ie width scaling) and the number of attention heads (\ie \# head scaling). Furthermore, we adapted AViT \cite{yin2022vit}, a dynamic token halting approach for image classification, to the 3D object detection task. We evaluate the latency of the backbone on a high-end NVIDIA A100 GPU and report the relative speed-up compared to the original architecture. Figure \ref{fig:pareto} plots the Pareto frontier of detection performance versus speed-up achieved by the various approaches. It is clear from the figure, that our method provides the best efficiency and accuracy trade-off. In fact, our method can achieve over a $50\%$ speed-up with only a slight impact to performance. Moreover, we find that AViT does not perform well, which indicates that a straightforward adaptation of a dynamic halting approach for image classification does not perform well when applied to the 3D object detection. We refer readers to Appendix \ref{apx:pareto} for more details on the experiment and a completed list of the numerical results.

\begin{table*}
\centering
\resizebox{\textwidth}{!}{
\begin{tabular}{lc|c|ll|ll|ll}
% \hline
\Xhline{2.1\arrayrulewidth}
\multirow{2}{*}{Method} & Publication & \multirow{2}{*}{TS} & \multicolumn{2}{c|}{Vehicle} & \multicolumn{2}{c|}{Pedestrian} & \multicolumn{2}{c}{Cyclist}\\
 & Date &  & AP/APH L1 & AP/APH L2 & AP/APH L1 & AP/APH L2 & AP/APH L1 & AP/APH L2\\\hline
PointPillar \cite{lang2019pointpillars} & CVPR 2019 & $\times$ & 72.1/71.5 & 63.6/63.1 & 70.6/56.7 & 62.8/50.3 & 64.4/62.3 & 61.9/59.9\\
PV-RCNN \cite{shi2020point} & CVPR 2020 & $\checkmark$ & 77.5/76.9 & 69.0/68.4 & 75.0/65.6 & 66.0/57.6  & 67.8/66.4 & 65.4/64.0\\
RangeDet \cite{fan2021rangedet} & ICCV 2021 & $\times$ & 72.9/72.3 & 64.0/63.6 & 75.9/71.9 & 67.6/63.9 & 65.7/64.4 & 63.3/62.1\\
MVP++ \cite{qi2021offboard} & CVPR 2021 & $\checkmark$ & 74.6/---~~~ & ~~~---/--- & 78.0/---~~~ & ~~~---/--- & ~~~---/--- & ~~~---/---\\
LiDAR R-CNN \cite{li2021lidar} & CVPR 2021 & $\checkmark$ & 76.0/75.5 & 68.3/67.9 & 71.2/58.7 & 63.1/51.7 & \underline{68.6/66.9} & \underline{66.1/64.4}\\
CenterPoint \cite{yin2021center} & CVPR 2021 & $\checkmark$ & 76.1/75.5  & 68.0/67.5 & 76.1/65.1 & 68.1/57.9 & ~~~---/--- & ~~~---/---\\
RSN \cite{sun2021rsn} & CVPR 2021 & $\times$ & 75.1/74.6 & 66.0/65.5 & 77.8/72.7 & 68.3/63.7 & ~~~---/--- & ~~~---/---\\
IA-SSD \cite{zhang2022not} & CVPR 2022 & $\checkmark$ & 70.5/69.7 & 61.6/61.0 & 69.4/58.5 & 60.3/50.7  & 67.7/65.3 & 65.0/62.7\\
LidarNAS \cite{liu2022lidarnas} & ECCV 2022 & $\times$ & 75.6/---~~~ & ~~~---/--- & 77.4/---~~~ & ~~~---/--- & ~~~---/--- & ~~~---/---\\

\hline 
% \Xhline{2.1\arrayrulewidth}
% VoTr-SSD \cite{mao2021voxel} & ICCV 2021 & $\times$ & 69.0/68.4 & 60.2/59.7 & ~~~---/--- & ~~~---/--- & ~~~---/--- & ~~~---/---\\
VoTr-TSD \cite{mao2021voxel} & ICCV 2021 & {\color{black}$\checkmark$} & 74.9/74.3 & 65.9/65.3 & ~~~---/--- & ~~~---/--- & ~~~---/--- & ~~~---/---\\
M3DETR \cite{guan2022m3detr} & WACV 2022 & $\times$ & 75.7/75.1 & 66.6/66.0  & 65.0/56.4 & 56.0/48.4 & 65.4/64.2 & 62.7/61.5\\
SWFormer \cite{sun2022swformer} & ECCV 2022 & $\times$ & \textbf{77.8/77.3} & \underline{69.2/68.8} & \underline{80.9/72.7} & \underline{72.5/64.9} & ~~~---/--- & ~~~---/---\\
SST \cite{fan2022embracing} & CVPR 2022 & $\times$ & 74.2/73.8 & 65.5/65.1 & 78.7/69.6 & 70.0/61.7 & ~~~---/--- & ~~~---/---\\
% \hdashline
\hline
SST$_{\text{center}}^*$ \cite{fan2022embracing} & CVPR 2022 & $\times$ & 76.2/75.7 & 67.7/67.2 & 79.9/71.4 & 72.7/64.8 & 67.7/66.3 & 65.2/63.8\\
SST\hspace{-0.1em}$\genfrac{}{}{0pt}{1}{++}{\text{halt}}$ (Ours) & - & $\times$ & \underline{77.7/77.1}\footnotesize{\color{blue}(+1.5)} & \textbf{69.5/69.0}\footnotesize{\color{blue}(+1.8)} & \textbf{80.9/73.0} \footnotesize{\color{blue}(+1.3)} & \textbf{74.0/66.5}\footnotesize{\color{blue}(+1.5)} & \textbf{70.0/68.6}\footnotesize{\color{blue}(+2.3)} & \textbf{67.3/66.0}\footnotesize{\color{blue}(+2.2)}\\[0.1em]
% \hline
\Xhline{2.1\arrayrulewidth}
% \bottomrule
\end{tabular}
}
\vspace{0.25em}
\caption{Performance comparison on the Waymo Open Dataset validation split. All methods take a single frame of LiDAR data as input. TS denotes whether or not a two-stage detection head is used. Methods below the first middle separator are transformer-based detectors. Note that, our proposed method, SST~\hspace{-0.4em}$\genfrac{}{}{0pt}{1}{++}{\text{halt}}$, does not apply any test-time augmentations, use a model ensemble, or use a two-stage detection head. $^*$The performance of SST$_{\text{center}}$ is based on our own implementation of SST with a CenterPoint detection head. The bold/underscored numbers correspond to the best and second best approach. The blue numbers in the parentheses are the average AP/APH improvement we obtain by applying our token halting approach to SST.}
\label{tbl:main}
\end{table*}

\begin{figure*}[t]
    \centering
    \includegraphics[width=0.95\textwidth]{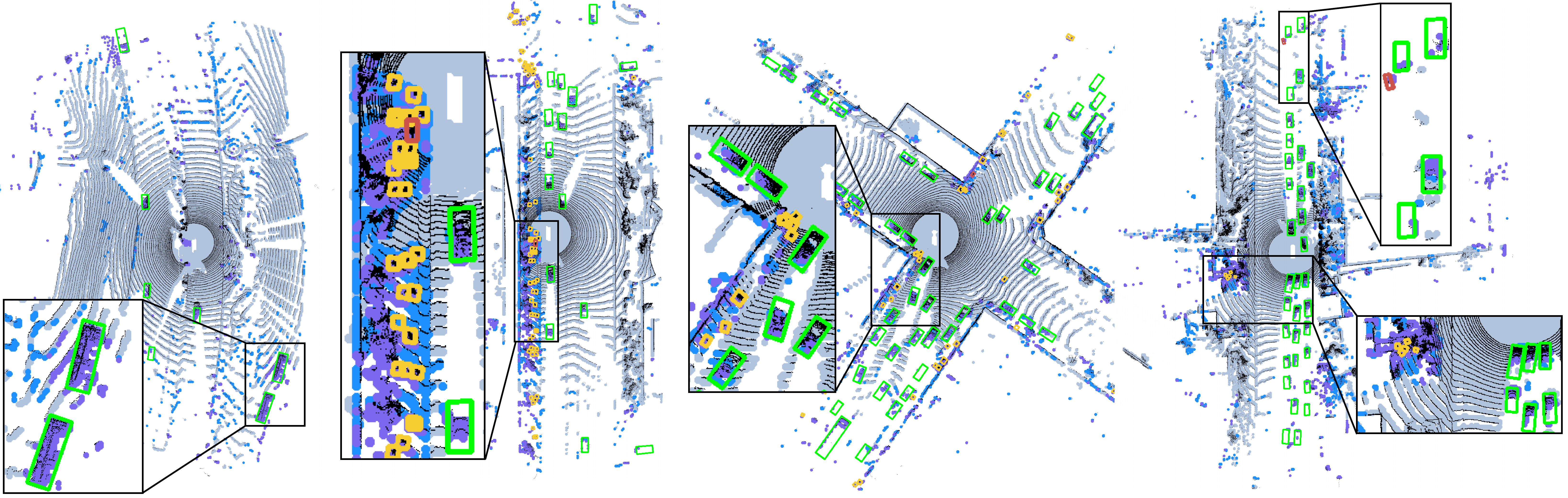}
    \vspace{0.25em}
    \caption{Visualization of the token halting process. The grey/blue voxels are halted at the first/second layer, and the purple voxels are not halted. The green, yellow, and red boxes are the vehicles, pedestrians, and cyclists detections, respectively.} 
    \label{fig:visual}
\end{figure*}

\subsection{Comparisons with the State-of-the-Art} \label{sec:sota}
In the previous section, we demonstrated that our proposed method can significantly reduce the latency of SST without dramatically impacting its performance.
The goal of this section is to leverage the latency savings provided by our dynamic token halting to improve the performance of SST while maintaining its runtime.
We improve the performance of SST by simply increasing the capacity of the detection head.
Specifically, we add an additional convolutional block and a feature pyramid network \cite{lin2017feature}.
The token sparsity is adjusted to ensure this improved SST has the same latency as the original model.
Refer to Appendix~\ref{apx:halt3d} for a detailed description of the architecture.

Table \ref{tbl:main} compares the performance of our improved model, referred to as SST~\hspace{-0.6em}$\genfrac{}{}{0pt}{1}{++}{\text{halt}}$, to the baseline model, SST~\cite{fan2022embracing}, on the Waymo Open Dataset validation split. We observe that the performance of our proposed method out-performs the baseline (SST). Compared to the original SST, we see a 1\% to 3.5\% AP/APH improvement for all classes. Recall that the improvement in detection performance is accomplished without increasing the latency of the model. Furthermore, our method meets or exceeds the performance of all other state-of-the-art models. Table \ref{tbl:test-set} shows the performance on the Waymo Open Data test split, and we observe a similar improvement.

% {\color{red} The performance of our improved SST model is similar to the recent SWFormer~\cite{sun2022swformer} but it is possible to also apply our method on SWFormer.}
% The performance of our improved SST model is similar to the recent SWFormer~\cite{sun2022swformer}; however, we want to emphasize that our dynamic token halting is a general framework that can be applied to any transformer-based backbone, including SWFormer,  to boost performance without sacrificing latency.

\begin{table}[t]
\centering
\resizebox{0.49\textwidth}{!}{
\begin{tabular}{l|ll|ll}
\Xhline{2.1\arrayrulewidth}
\multirow{2}{*}{Method} & \multicolumn{2}{c|}{Vehicle} & \multicolumn{2}{c}{Pedestrian} \\
  & AP/APH L1 & AP/APH L2 & AP/APH L1 & AP/APH L2 \\
\hline
PointPillar & 68.6/68.1 & 60.5/60.1 & 68.0/55.5 & 61.4/50.1\\
CenterPoint\_2f & 80.2/79.7 & 72.2/71.8 & 78.3/72.1 & 72.2/66.4\\
\hline
SST$_{\text{center}}$ & 80.0/79.6 & 72.1/71.7 & 79.3/71.3 & 73.4/65.9\\
SST\hspace{-0.1em}$\genfrac{}{}{0pt}{1}{++}{\text{halt}}$ (Ours) & \textbf{81.2/80.7} \footnotesize{\color{blue}(+1.2)} & \textbf{73.7/73.2} \footnotesize{\color{blue}(+1.6)} & \textbf{80.4/72.8} \footnotesize{\color{blue}(+1.3)} & \textbf{74.7/67.5} \footnotesize{\color{blue}(+1.5)} \\[0.1em]
\Xhline{2.1\arrayrulewidth}
\end{tabular}
}
\vspace{0.25em}
\caption{Performance on the Waymo Open Dataset test split. The setting is the same as Table \ref{tbl:main}.} \label{tbl:test-set}
\end{table}

\begin{table*}[t]
\centering{}%
\resizebox{0.75\textwidth}{!}{
\begin{tabular}{l|ll|ll|ll}
% \toprule
% \specialrule{0.1em}{.07em}{.07em}
% \hline
\Xhline{2.1\arrayrulewidth}
\multirow{2}{*}{Config} & \multicolumn{6}{c}{Vehicle BEV AP}\tabularnewline
 & {[}0, 30) L1 & {[}0, 30) L2 & {[}30, 50) L1 & {[}30, 50) L2 & {[}50, $\infty$) L1 & {[}50, $\infty$) L2\\
\hline 
SST$_{\text{center}}$ & 92.2 & 91.0 & 74.6 & 68.0 & 52.7 & 40.8\\
SST$_{\text{halt}}$ & 92.2 \small{\color{gray}(+0.0)} & 91.1 \small{\color{blue}(+0.1)} & 74.6 \small{\color{gray}(+0.0)} & 68.5 \small{\color{blue}(+0.5)} & 52.7 \small{\color{gray}(+0.0)} & 41.2 \small{\color{blue}(+0.4)}\\
SST\hspace{-0.1em}$\genfrac{}{}{0pt}{1}{++}{\text{halt}}$ & 92.6 \small{\color{blue}(+0.4)} & 91.4 \small{\color{blue}(+0.3)} & 76.2 \small{\color{blue}(+1.6)} & 70.0 \small{\color{blue}(+2.0)} & 55.4 \small{\color{blue}(+2.7)} & 43.4 \small{\color{blue}(+2.6)}\\[0.1em]
% \bottomrule 
% \hline
\Xhline{2.1\arrayrulewidth}
\end{tabular}
}
\vspace{0.5em}
\caption{Breakdown of the BEV vehicle detection performance. SST$_{\text{center}}$ is our implementation of SST with a CenterPoint detection head. SST$_{\text{halt}}$ is the SST model with our dynamic halting (the model explored in section \ref{sec:pareto}). SST\hspace{-0.1em}$\genfrac{}{}{0pt}{1}{++}{\text{halt}}$ is the model we proposed in section \ref{sec:sota}. The blue numbers in the parentheses are the improvement we obtained compared with SST$_{\text{center}}$.} \label{tbl:decompose}
\end{table*}

\subsection{Additional Analysis}
We find that dynamic token halting is not only an effective approach to obtain a better efficiency-accuracy trade-off, but also it has the additional effect of aiding in the detection of long-range and difficult objects.
In Table \ref{tbl:decompose}, we show the breakdown of detection performance at different distances and difficulty levels for the original SST, the SST with dynamic token halting (denoted as SST$_{\text{halt}}$), and our improved SST\hspace{-0.1em}$\genfrac{}{}{0pt}{1}{++}{\text{halt}}$. Compared to the baseline, SST$_{\text{halt}}$ has better accuracy for the difficult L2 objects while the same performance for L1 objects. Moreover, for both SST$_{\text{halt}}$ and SST\hspace{-0.1em}$\genfrac{}{}{0pt}{1}{++}{\text{halt}}$, the improvement becomes more significant for long-range objects. We believe this result can be explained intuitively. For difficult and long-range objects, their points are more sparse, which may make it challenging for the model to distinguish the foreground objects from the background. Our dynamic token halting method removes the majority of the background tokens, which perhaps increases the signal-to-noise ratio for the detector.

\begin{table*}[t]
\centering{}%
\resizebox{0.99\textwidth}{!}{
\begin{tabular}{l|ll|ll|ll}
% \toprule 
% \specialrule{0.1em}{.07em}{.07em}
% \hline
\Xhline{2.1\arrayrulewidth}
\multirow{2}{*}{Config} & \multicolumn{2}{c|}{Vehicle} & \multicolumn{2}{c|}{Pedestrian} & \multicolumn{2}{c}{Cyclist}\tabularnewline
 & AP/APH L1 & AP/APH L2 & AP/APH L1 & AP/APH L2 & AP/APH L1 & AP/APH L2\tabularnewline
\hline 
Full & 75.7/75.2 & 67.7/67.2 & 78.7/70.4 & 72.3/64.5 & 70.0/68.5 & 67.8/66.2\tabularnewline
w/o voxel recycle & 61.9/61.0 \small{\color{red}(-14.0)} & 54.9/54.0 \small{\color{red}(-13.0)} & 76.8/68.8 \small{\color{red}(-1.8)} & 69.7/62.2 \small{\color{red}(-2.5)} & 63.7/62.3 \small{\color{red}(-6.2)} & 61.3/60.0 \small{\color{red}(-6.4)} \tabularnewline
w/o non-uniform sparsity & 74.5/74.0 \small{\color{red}(-1.2)} & 66.1/65.6 \small{\color{red}(-1.6)} & 77.7/68.1 \small{\color{red}(-1.6)} & 70.2/61.3 \small{\color{red}(-2.6)} & 64.2/62.7 \small{\color{red}(-5.8)} & 61.8/60.3 \small{\color{red}(-6.0)} \tabularnewline
% \bottomrule 
% \hline
\Xhline{2.1\arrayrulewidth}
\end{tabular}
}
\vspace{0.5em}
\caption{Ablation study where ``w/o non-uniform sparsity'' denotes that we use a uniform sparsity penalty instead. The red numbers in the parentheses are the average AP/APH decrease when a certain component is removed.}\label{tbl:ablation}
\end{table*}

\subsection{Ablation Study}
We conducted an ablation study to show the efficacy of the various aspects of our proposed method. For each experiment, we keep the sparsity/latency the same and study how the performance changes when we remove different components.
Specifically, we are interested in how the performance changes when the token recycling is removed and  when the non-uniform sparsity loss is replaced with an uniform sparsity loss. Table \ref{tbl:ablation} summarizes the result. 

We observe a considerable drop in performance, especially for vehicles and cyclists, when we do not recycle the halted tokens. We do not believe this is because the halting module is not keeping important tokens. If it was not capturing important tokens, we would have seen a significant drop in performance even with token recycling because when a token is halted, it will only be processed by one or two attention blocks which is only a quarter or half of the backbone. However, in Section \ref{sec:pareto}, there is very little performance drop even with very high sparsity. We believe the performance degradation is due to the halted tokens still having useful semantic information for the detection and thus should be used by the detection head. 

% Also, it is inevitable that some tokens may be mistakenly halted early and the recycling mechanism enables the network to be more tolerant of such mistakes.

\begin{figure}[t]
    \centering
    \includegraphics[width=0.235\textwidth]{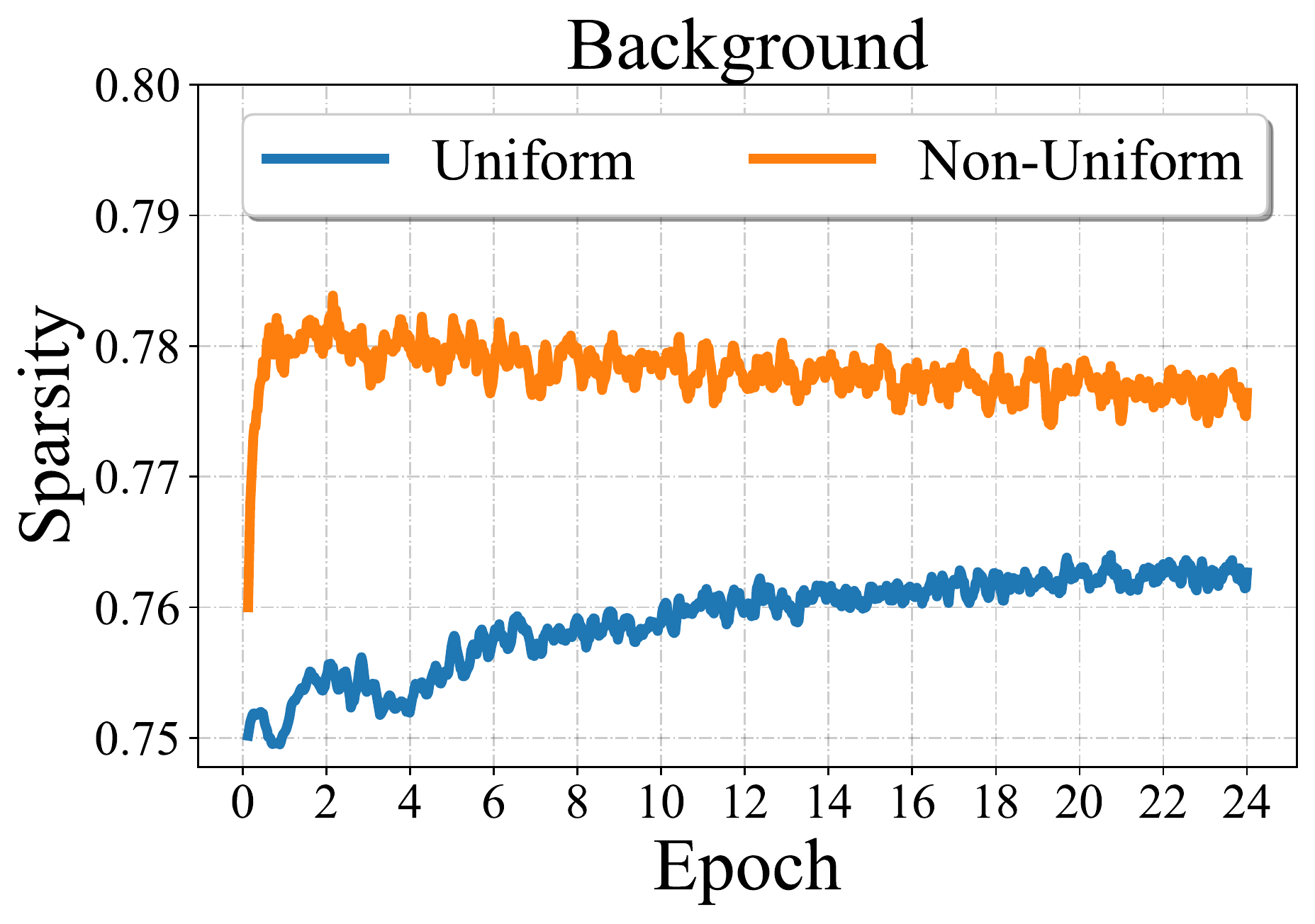}
    \includegraphics[width=0.235\textwidth]{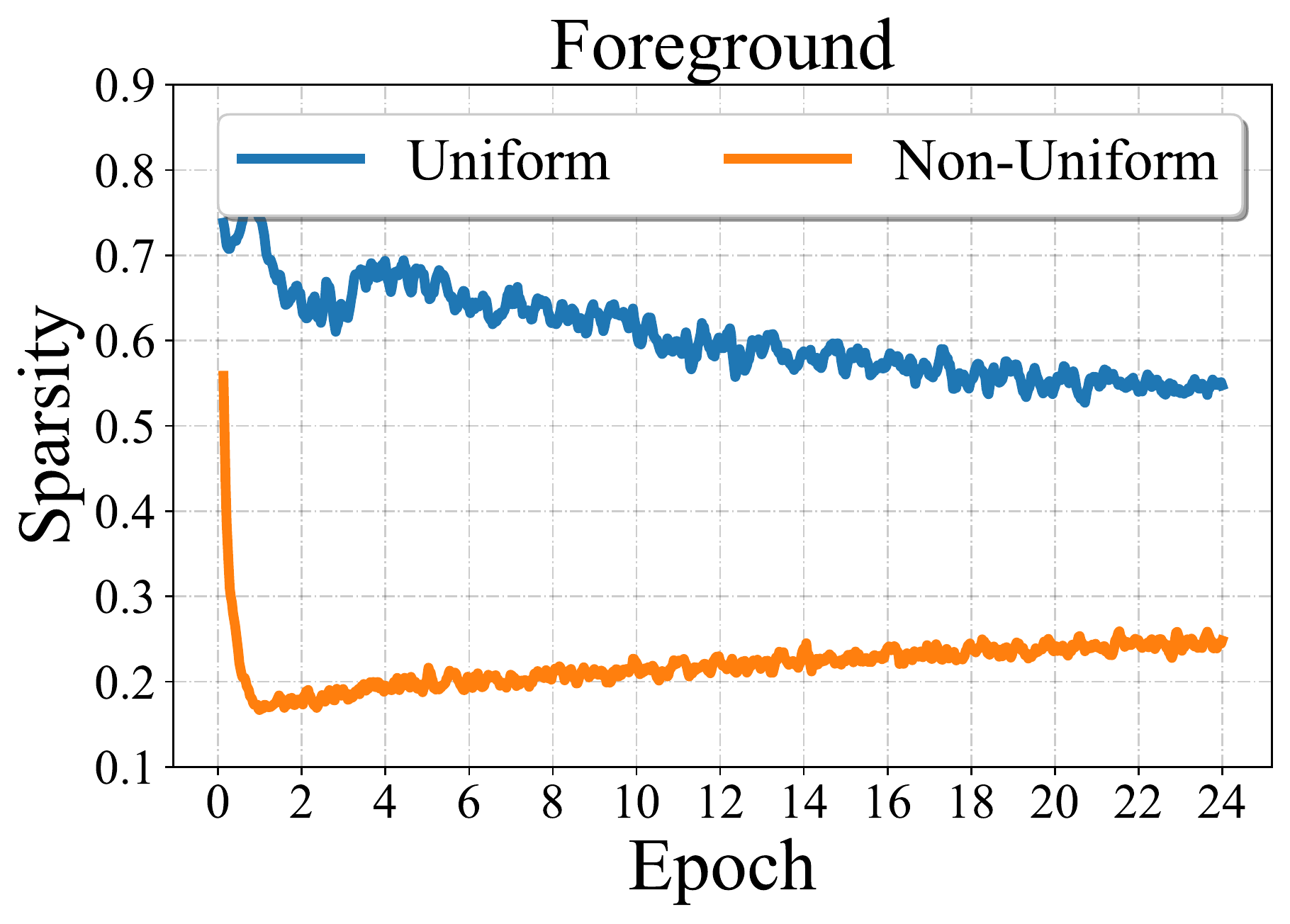}
    \caption{The sparsity of background and foreground tokens when training with the uniform and non-uniform sparsity loss.} 
    \label{fig:sparsity}
\end{figure}

The non-uniform sparsity loss is also shown to be useful. To demonstrate this, we replaced the non-uniform sparsity loss with a uniform sparsity loss, \ie $\ell_1$ loss. Since our framework is fully differentiable, the model is still able to achieve good results with a uniform sparsity loss. However, the non-uniform loss provides a better signal to the model. To understand how, we plot the change in sparsity of background and foreground tokens when training with and without our non-uniform sparsity loss in Figure \ref{fig:sparsity}.

With the uniform sparsity loss, as training progresses, the ratio of kept foreground tokens increases while the ratio of background tokens decreases. We believe this occurs because the model gradually becomes better at the detection task and in turn becomes better at selecting useful tokens.
However, during the early stages of training, the model fails to make a good token halting decision, which negatively impacts the learning of the model. As a result, the model ends up in a sub-optimal state. When the non-uniform sparsity loss is applied, we observe two interesting phases during learning. In the first phase (within the first epoch), the model is learning to select tokens primarily based on whether they belong to a foreground object. In this phase, the sparsity of foreground tokens rapidly decreases while the sparsity of background tokens rapidly increases. Afterwards, the model goes into a second learning phase, and it starts to adjust the selection of tokens based on their contribution to the detection task. In this phase, the model learns to halt useless foreground tokens while keeping useful background tokens.

\subsection{Visualization}
To better understand the token halting process, we visualize the tokens halted at different layers with different colors in Figure \ref{fig:visual}. From the figure, we observe that most of the tokens belong to roads, sides of buildings, or foliage, which do not contribute much to the detection. Appropriately, these tokens are halted at the very beginning by the model. In addition, our model tends to keep tokens that are near the objects we want to detect, since those tokens are usually semantically informative. For example, in the leftmost portion of the second scene, many of the background voxels near pedestrians are kept while the voxels in the rightmost portion are almost all halted as there are no pedestrians. Lastly, we observed that tokens belonging to objects remained un-halted with a significantly higher chance as those tokens are in general critical for the detection.

% \begin{figure*}[t]
%     \centering
%     \includegraphics[width=0.35\textwidth]{fig/visual/s29.png}
%     \hspace{-0.7cm}
%     \includegraphics[width=0.35\textwidth]{fig/visual/s24.png}
%     \hspace{-0.7cm}
%     \includegraphics[width=0.35\textwidth]{fig/visual/s22.png}
%     \includegraphics[width=0.35\textwidth]{fig/visual/s25.png}
%     \hspace{-0.7cm}
%     \includegraphics[width=0.35\textwidth]{fig/visual/s21.png}
%     \hspace{-0.7cm}
%     \includegraphics[width=0.35\textwidth]{fig/visual/s9.png}
%     \caption{Visualization of the voxel halting process. The grey/blue voxels are halted at the first/second layer and the purple voxels are not halted during the forward computation. The green/yellow/red boxes are vehicles/pedestrians/cyclists.} 
%     \label{fig:visual}
% \end{figure*}

\section{Conclusion}
In this work, we proposed an approach to dynamically halt tokens in order to speed-up transformer-based 3D object detectors. Our method significantly improves the Pareto frontier of the accuracy and efficiency trade-off. By leveraging our improved model efficiency, we are able to dramatically increase the performance of the baseline model without sacrificing latency. There are several interesting directions for future research. We believe there are additional architectural changes that could be made to further exploit the high-level of token sparsity and to further reduce model latency. It would also be interesting to explore how to generalize this idea to multi-modal and multi-frame object detectors.

{\small
\bibliographystyle{ieee_fullname}
\bibliography{egbib}
}

\appendix
\onecolumn
\section{Appendix}

\subsection{Additional Details}

\subsubsection{Non-Uniform Sparsity Loss} \label{apx:non-unif}
For each token, we determine whether or not it lies within a ground-truth bounding box.
If the $i$-th token falls inside a bounding box, the corresponding heatmap's value is defined as
\begin{equation}
    m_i =\exp\left(-\frac{(x_i-l_{x})^{2}+(y_i-l_{y})^{2}}{2\sigma^{2}}\right)\text{,}
\end{equation}
where $x_i$ and $y_i$ are the $xy$-coordindates of the token, $l_x$ and $l_y$ are the $xy$-coordinates of the box's center, and $\sigma$ is a hyper-parameter that controls the smoothness of the heatmap. For tokens that do not lie within a bounding box, the heatmap's value is set to zero. We create the heatmap for each object class, and we take the maximize over all heatmaps to obtain the final class-agnostic heatmap used by our non-uniform sparsity loss. The non-uniform sparsity loss is similar to focal loss \cite{zhou2019objects}, and it is defined as follows:
\begin{equation}
\L_{s} = -\sum_{l=1}^{L}\sum_{i \in \mathcal{K}}\frac{1}{|\mathcal{K}|}\left[(1-s_{l,i})^{\alpha}\log(s_{l,i})\mathbb{I}_{m_{i}\ge1-\epsilon} +(1-\hm_{i})^{\gamma}s_{i}^{\alpha}\log(1-s_{l,i})\mathbb{I}_{m_{i}<1-\epsilon}\right]\text{,}
\end{equation}
where $\mathcal{K} = \{i:k_{0:l-1, i}=1\}$ is the set of tokens that have not been halted before the $l$-th layer, $s_{l,i}$ is the score for the $i$-th token at the $l$-th layer, $\alpha=2$ and $\gamma=4$ are hyper-parameters, and $\epsilon=10^{-4}$ is used to improve numerical stability.

\subsubsection{Analyzing the Pseudo-Gradient} 
\label{apx:gradient}

In Section \ref{sec:grad}, we claim the following:
\begin{equation}\label{equ:approx1}
    \Delta_{i}\approx \frac{\partial\L(q_{1},q_{2})}{\partial{s_{i}}} + O(u)\text{,}
\end{equation}
where
\begin{equation}
    \Delta_{i}:=\L(\tilde{q}_{1},\tilde{q}_{2})-\L(q_{1},q_{2})
\end{equation} 
is the difference in the detection loss when the $i$-th token is halted instead of being forwarded in a single layer network.
In other words, we claim that the (pseudo)-gradient of $\L(\tilde{q}_{1},\tilde{q}_{2})$ with respect to $s_i$ provided by our proposed EDF and the STE is a reasonable proxy of $\Delta_{i}$.

To prove this claim, we begin by computing
\begin{equation}
    \frac{\partial\L(q_{1},q_{2})}{\partial{s_{i}}}=\left\langle \frac{\partial\L(q_{1},q_{2})}{\partial{q_{1}}},\frac{\partial{q_{1}}}{\partial{s_{i}}}\right\rangle +\left\langle \frac{\partial\L(q_{1},q_{2})}{\partial{q_{2}}},\frac{\partial{q_{2}}}{\partial{s_{i}}}\right\rangle\text{.}
\end{equation}
Recall that $q_{1} =(\1-k)\circ f$ and $q_{2}=k\circ\phi_{2}(\wa(\phi_{1}(f),s\circ k),f)$.
Using the definition of the STE, 
\begin{equation}
    \frac{\partial q_{1}}{\partial s_i} = -\frac{\partial(k\circ f)}{\partial s_i} = -\1_{i}\circ f
    \label{eq:grad-1}
\end{equation}
and
\begin{align}
    \frac{\partial q_{2}}{\partial s_i} &= k\circ \frac{\partial \phi_{2}(\wa(\phi_{1}(f),s\circ k),f)}{\partial s_{i}} + \frac{\partial k}{\partial s_{i}} \circ\phi_{2}(\wa(\phi_{1}(f),s\circ k),f)\\
    &= k\circ \frac{\partial \phi_{2}(\wa(\phi_{1}(f),s\circ k),f)}{\partial s_{i}} + \1_{i}\circ\phi_{2}(\wa(\phi_{1}(f),s\circ k),f)\text{.}
    \label{eq:grad-2}
\end{align}
Next, let us compute $\Delta_{i}$.
Using Taylor series approximation,
\begin{equation}
    \Delta_{i}\approx\left\langle \frac{\partial\L(q_{1},q_{2})}{\partial{q_{1}}},\tilde{q}_{1}-q_{1}\right\rangle +\left\langle \frac{\partial\L(q_{1},q_{2})}{\partial{q_{2}}},\tilde{q}_{2}-q_{2}\right\rangle\text{.}
\end{equation}
Recall that $\tilde{q}_{1} =(\1-k-\textbf{\ensuremath{\1}}_{i})\circ f$ and $\tilde{q}_{2} =(k+\1_{i})\circ\phi_{2}(\wa(\phi_{1}(f),s\circ(k+\1_{i})),f)$; therefore,
\begin{equation}
    \tilde{q}_{1}-q_{1} = -\1_{i}\circ f
    \label{eq:grad-3}
\end{equation}
and
\begin{align}
\tilde{q}_{2}-q_{2} & =(k+\1_{i})\circ\phi_{2}(\wa(\phi_{1}(f),s\circ(k+\1_{i})),f)-k\circ\phi_{2}(\wa(\phi_{1}(f),s\circ k),f)\\
 & =k\circ\phi_{2}(\wa(\phi_{1}(f),s\circ(k+\1_{i})),f)-k\circ\phi_{2}(\wa(\phi_{1}(f),s\circ k),f)\\
 & +\1_{i}\circ\phi_{2}(\wa(\phi_{1}(f),s\circ(k+\1_{i})),f)\text{.}
\end{align}
Again, using Taylor series approximation, 
\begin{equation}
k\circ\phi_{2}(\wa(\phi_{1}(f),s\circ(k+\1_{i})),f)-k\circ\phi_{2}(\wa(\phi_{1}(f),s\circ k),f) \approx  k\circ \frac{\partial \phi_{2}(\wa(\phi_{1}(f),s\circ k),f)}{\partial s_{i}} 
\end{equation}
and
\begin{equation}
    \1_{i}\circ\phi_{2}(\wa(\phi_{1}(f),s\circ(k+\1_{i})),f) \approx \1_{i}\circ\phi_{2}(\wa(\phi_{1}(f),s\circ k),f) + \1_{i}\circ \frac{\partial \phi_{2}(\wa(\phi_{1}(f),s\circ k),f)}{\partial s_{i}} 
\end{equation}
As a result,
\begin{equation}
\tilde{q}_{2}-q_{2} \approx k\circ \frac{\partial \phi_{2}(\wa(\phi_{1}(f),s\circ k),f)}{\partial s_{i}} + \1_{i}\circ\phi_{2}(\wa(\phi_{1}(f),s\circ k),f) + \1_{i}\circ \frac{\partial \phi_{2}(\wa(\phi_{1}(f),s\circ k),f)}{\partial s_{i}}\text{.}
\label{eq:grad-4}
\end{equation}
Comparing Eq. \eqref{eq:grad-1} to Eq. \eqref{eq:grad-3} and Eq. \eqref{eq:grad-2} to Eq. \eqref{eq:grad-4}, we see that
\begin{align}
    \Delta_{i} - \frac{\partial\L(q_{1},q_{2})}{\partial{s_{i}}} &\approx \left\langle \frac{\partial\L(q_{1},q_{2})}{\partial{q_{2}}},\1_{i}\circ \frac{\partial \phi_{2}(\wa(\phi_{1}(f),s\circ k),f)}{\partial s_{i}}\right\rangle \\
    &= \frac{\partial\L(q_{1},q_{2})}{\partial{q_{2, i}}} \frac{\partial \phi_{2}(\wa(\phi_{1}(f),s\circ k),f)_i}{\partial s_{i}} \\
    &= \frac{\partial\L(q_{1},q_{2})}{\partial{q_{2, i}}} \frac{\partial \phi_{2}(\wa(\phi_{1}(f),s\circ k),f)_i}{\partial \wa(\phi_{1}(f),s\circ k)_i} \frac{\partial \wa(\phi_{1}(f),s\circ k)_i}{\partial (s \circ k)_i} s_i \text{,}
\end{align}
where
\begin{equation}
    \frac{\partial \phi_{2}(\wa(\phi_{1}(f),s\circ k),f)_i}{\partial s_{i}} = \frac{\partial \phi_{2}(\wa(\phi_{1}(f),s\circ k),f)_i}{\partial \wa(\phi_{1}(f),s\circ k)_i} \frac{\partial \wa(\phi_{1}(f),s\circ k)_i}{\partial (s \circ k)_i} \frac{\partial (s \circ k)_i}{\partial s_i}
\end{equation}
and
\begin{equation}
    \frac{\partial (s \circ k)_i}{\partial s_i} = \frac{\partial s_i k_i}{\partial s_i} = k_i + s_i = s_i
\end{equation}
using the definition of the STE and the fact that $k_i = 0$.
In general, the derivative of $\L$ and $\phi_{2}$ is bounded since the parameter space of the network is bounded (due to the weight decay) and the operators inside the network are Lipschitz continuous.
However, $\partial \wa(\phi_{1}(f),s\circ k)_i / \partial (s \circ k)_i$ can be singular when all the element of $s\circ k$ are zero. We argue that in our analysis, we can still treat this term as bounded for two reasons. Firstly, in practice, we add an $\epsilon$ to the denominator of the $\wa$ to prevent numeric instability, which makes the gradient bounded even if all the elements of $s\circ k$ are zero. Secondly, we employ gradient clipping to enforce a bound on the gradient.
All of this combine, we have 
\begin{equation}
    \Delta_{i}\approx \frac{\partial\L(q_{1},q_{2})}{\partial{s_{i}}}+ O(u)\text{.}
\end{equation}
That is, the approximation error of the pseudo-gradient is proportional to $s_{i}$ thanks
to the usage of weighted attention. Furthermore, since $k_{i}=0$, we have $\left|s_{i}\right|<u$ where the threshold $u$ is in general a very small value. This demonstrates that our pseudo-gradient provides useful information for updating the halting module.

\subsection{Additional Experiment Details and Results}

\subsubsection{Setup and Implementation Details} \label{apx:setup}
We use a mixed strategy to decide the halting threshold. We specific an upper and lower token score quantile (denoted as $\alpha_u$ and $\alpha_l$) and enforce that the sparsity of each layer varies within $[\alpha_l, \alpha_u]$. To achieve this, the final threshold is given by clamping the pre-specific threshold $u$ within $[Q(\alpha_l), Q(\alpha_u)]$ where $Q(\alpha_l)$ and $Q(\alpha_u)$ denote the score corresponding to the $\alpha_l$ and $\alpha_u$ quantile, respectively. We enforce such a constrain because we observe that the distribution of scores can vary considerably for different scenes during the early stages of training and selecting the threshold in this way helps to stabilize training. In Table \ref{tbl:main}, the sparsity is bounded between 80\% and 90\% for the first halting module, 90\% and 99\% for the second halting module, and the default value of $u$ is 0.01. The following technique can be used to identify $u$ for a new dataset/model: train a model for a short period, then select $u$ such that it is higher than the score of most foreground voxels and less than the score of most background voxels. 

% Our U-Net architecture follows \cite{ronneberger2015u}. We downsample using four MobileNet V2 \cite{sandler2018mobilenetv2} convolution blocks with stride 2. Expect for the first block that takes 32 input channel and outputs 16 channels, for the other blocks, the number of output channels is 2 times the number of input channels. After that we use four bi-linear upsampling layers.

\subsubsection{Efficiency and Accuracy Trade-off} \label{apx:pareto}
For the baselines, we vary the latent dimension of the attention mechanism by $\{16, 12, 8, 4\}$, and we vary the number of attention head by $\{8, 6, 5, 4\}$. We adapt AViT from \cite{yin2022vit}, but we apply our token recycling to improve the performance. Also, we adjust the number of input token features for the halting module from 1 to 32 as this improves performance while having a negligible impact on latency.
Table \ref{tbl:pareto} summarizes the results. Overall, we observe that our method significantly improves over other model scaling approaches as well as AViT.

\begin{table*}[t]
\centering{}%
\resizebox{\textwidth}{!}{
\begin{tabular}{c|c|cc|cc|cc}
\Xhline{2.1\arrayrulewidth}
\multirow{2}{*}{Speed Up/Sparsity} & \multirow{2}{*}{Method} & \multicolumn{2}{c|}{Vehicle} & \multicolumn{2}{c|}{Pedestrian} & \multicolumn{2}{c}{Cyclist}\tabularnewline
 &  & AP/APH L1 & AP/APH L2 & AP/APH L1 & AP/APH L2 & AP/APH L1 & AP/APH L2\tabularnewline
\hline 
1.00/0.00 & Original & 76.2/75.7 & 67.7/67.2 & 79.9/71.4 & 72.7/64.8 & 67.7/66.3 & 65.2/63.8\tabularnewline
\hline 
1.07/0.00 & Width scale & 75.4/74.9 & 66.9/66.5 & 79.5/70.7 & 72.2/64.0 & 65.6/64.1 & 63.0/61.6\tabularnewline
1.16/0.00 & Num head scale & 75.5/75.0 & 67.0/66.6 & 79.4/70.8 & 72.0/64.1 & 65.5/63.9 & 63.0/61.5\tabularnewline
1.28/0.48 & AViT$_{\text{adapted}}$ \cite{yin2022vit} & 71.9/71.3 & 63.4/62.9 & 76.8/67.9 & 69.1/60.9 & 63.1/61.6 & 60.7/59.3 \tabularnewline
1.21/0.75 & Ours & 76.1/75.6 & 67.8/67.3 & 79.4/70.7 & 72.1/64.0 & 67.0/65.6 & 64.4/63.1\tabularnewline
\hline 
1.13/0.00 & Width scale & 73.8/73.3 & 65.3/64.8 & 78.2/69.0 & 70.6/62.1 & 61.7/60.2 & 59.3/57.9\tabularnewline
1.45/0.00 & Num head scale & 73.8/73.3 & 65.4/64.9 & 78.2/68.8 & 70.7/62.1 & 62.0/60.3 & 59.6/58.0\tabularnewline
1.39/0.57 & AViT$_{\text{adapted}}$ \cite{yin2022vit} & 70.3/69.7 & 61.9/61.4 & 76.2/67.2 & 68.4/60.1 & 60.8/59.3 & 58.5/57.0\tabularnewline
1.37/0.82 & Ours & 76.1/75.6 & 67.7/67.2 & 79.9/71.5 & 72.6/64.7 & 67.4/66.1 & 64.8/63.6\tabularnewline
\hline 
1.18/0.00 & Width scale & 70.4/69.8 & 62.0/61.5 & 74.5/64.3 & 66.7/57.4 & 54.9/52.9 & 52.8/50.8\tabularnewline
% 1.81/ --- & Num head scale & 70.5/70.0 & 62.1/61.6 & 74.3/63.9 & 66.5/57.0 & 55.2/53.1 & 53.1/51.1\tabularnewline
1.45/0.00 & Num head scale & 73.8/73.3 & 65.4/64.9 & 78.2/68.8 & 70.7/62.1 & 62.0/60.3 & 59.6/58.0\tabularnewline
1.55/0.72 & AViT$_{\text{adapted}}$ \cite{yin2022vit} & 70.1/69.6 & 61.7/61.3 & 76.3/67.5 & 68.5/60.4 & 61.6/60.1 & 59.2/57.8\tabularnewline
1.52/0.89 & Ours & 75.4/74.9 & 67.0/66.5 & 79.7/71.5 & 72.4/64.7 & 67.1/65.7 & 64.5/63.2\tabularnewline
\Xhline{2.1\arrayrulewidth}
\end{tabular}
}
\caption{Efficiency and accuracy trade-off. We report the relative backbone speed-up and the average sparsity across all the attention layers.} \label{tbl:pareto}
\end{table*}

\subsubsection{The SST\hspace{-0.1em}$\genfrac{}{}{0pt}{1}{++}{\text{halt}}$ Architecture} \label{apx:halt3d}
For our SST\hspace{-0.1em}$\genfrac{}{}{0pt}{1}{++}{\text{halt}}$ architecture, we use a U-Net \cite{ronneberger2015u} and a single layer MLP as the first and second halting module. We find that the latent features of the U-Net contain useful semantic information. To reuse those features, we fuse the token features with the U-Net's features by applying a linear transformation and sums the features. Furthermore, we add the U-Net's feature map to the BEV feature map. To leverage the latency savings provided by halting tokens, SST\hspace{-0.1em}$\genfrac{}{}{0pt}{1}{++}{\text{halt}}$ uses an extra convolutional block in the detection head for a total of two convolutional blocks. The first convolutional block contains four convolutional layers. The second convolutional block contains four convolutional layer where the first layer has a stride of 2. All the layers use a kernel size of 3. Afterwards, we use the Feature Pyramid Network \cite{lin2017feature} employed by SECOND \cite{yan2018second} to fuse the two scales of the BEV feature map and make predictions.

% \begin{table*}
% \centering{}%
% \begin{tabular}{l|c|cc|cc}
% \toprule 
% \multirow{2}{*}{Method} & \multirow{2}{*}{TS/\#f} & \multicolumn{2}{c|}{Vehicle} & \multicolumn{2}{c}{Pedestrian}\tabularnewline
%  &  & AP/APH L1 & AP/APH L2 & AP/APH L1 & AP/APH L2\tabularnewline
% \hline 
% CenterPoint & $\checkmark/1$f & 80.2/79.7  & 72.2/71.8 & 78.3/72.1 & 72.2/66.4\tabularnewline
% PVRCNN++ & $\checkmark/1$f & 81.6/81.2 & 73.9/73.5 & 80.4/75.0 & 74.1/69.0\tabularnewline
% PointPillars & $\times/1$f & 68.6/68.1 & 60.5/60.1 & 68.0/55.5 & 61.4/50.1\tabularnewline
% RSN\_3f & $\times/3$f & 80.7/80.3 & 71.9/71.6 & 78.9/75.6 & 70.7/67.8\tabularnewline
% SWFormer\_3f & $\times/3$f & 82.9/82.5  & 75.0/74.6 & 82.1/78.1 & 75.9/72.1\tabularnewline
% SST\_TS\_3f & $\checkmark/3$f & 81.0/80.6 & 73.1/72.7 & 83.0/79.4 & 76.6/73.1\tabularnewline
% \hline 
% SST$_\text{center}$\_1f & $\checkmark/1$f & 80.0/79.6 & 72.1/71.7 & 79.3/71.3 & 73.4/65.9 \tabularnewline
% SST$_\text{center}$\_3f & $\times/3$f & 81.3/80.9 & 73.8/73.4 & 78.3/72.9 & 72.8/67.6 \tabularnewline
% Halt3d & $\times/1$f & 81.2/80.7 & 73.7/73.2 & 80.4/72.8 & 74.7/67.5\tabularnewline
% Halt3d\_3f & $\times/3$f & 81.8/81.3 & 74.6/74.2 & 79.2/73.8 & 73.6/68.6\tabularnewline
% \bottomrule 
% \end{tabular}\caption{WOD test}
% \end{table*}

\end{document}